\documentclass[journal]{IEEEtran}

\usepackage{times}
\usepackage{mathrsfs}
\usepackage{epsfig}
\usepackage{graphicx}
\usepackage{amsmath}
\usepackage{amssymb}
\usepackage{graphicx}
\usepackage{cite}
\usepackage{enumerate}
\usepackage{cases}
\usepackage{multirow}
\usepackage{verbatim}
\usepackage{amssymb}
\usepackage{CJK}
\usepackage{algorithm}
\usepackage{algorithmicx}
\usepackage{algpseudocode}
\usepackage{color}
\usepackage{bm}
\usepackage{booktabs}
\usepackage{amssymb}
\usepackage[colorlinks,linkcolor=red]{hyperref}

\newcommand{\etal}{\textit{et al}.}
\newcommand{\ie}{\textit{i}.\textit{e}.}
\newcommand{\eg}{\textit{e}.\textit{g}.}
\newcommand{\etc}{\textit{etc}}

\begin{document}

\title{Global-and-Local Collaborative Learning for Co-Salient Object Detection}

\author
{
Runmin Cong,~\IEEEmembership{Member,~IEEE,} Ning Yang, Chongyi Li, Huazhu Fu, Yao Zhao,~\IEEEmembership{Senior Member,~IEEE,}\\ Qingming Huang,~\IEEEmembership{Fellow,~IEEE,} and Sam Kwong,~\IEEEmembership{Fellow,~IEEE}
\thanks{Runmin Cong is with the Institute of Information Science, Beijing Jiaotong University, Beijing 100044, China, also with the Beijing Key Laboratory of Advanced Information Science and Network Technology, Beijing 100044, China, and also with the Department of Computer Science, City University of Hong Kong, Hong Kong SAR, China (e-mail: rmcong@bjtu.edu.cn).}
\thanks{Ning Yang and Yao Zhao are with the Institute of Information Science, Beijing Jiaotong University, Beijing 100044, China, and also with the Beijing Key Laboratory of Advanced Information Science and Network Technology, Beijing 100044, China (e-mail: ningyang@bjtu.edu.cn; yzhao@bjtu.edu.cn).}
\thanks{Chongyi Li is with the  School of Computer Science and Engineering, Nanyang Technological University, Singapore (e-mail: lichongyi25@gmail.com).}
\thanks{Huazhu Fu is with Institute of High Performance Computing (IHPC), A*STAR, Singapore (e-mail: hzfu@ieee.org).}
\thanks{Qingming Huang is with the School of Computer Science and Technology, University of Chinese Academy of Sciences, Beijing 101408, China, also with the Key Laboratory of Big Data Mining and Knowledge Management (BDKM), University of Chinese Academy of Sciences, Beijing 101408, China, also with the Key Laboratory of Intelligent Information Processing, Institute of Computing Technology, Chinese Academy of Sciences, Beijing 100190, China, and also with Peng Cheng Laboratory, Shenzhen 518055, China (email: qmhuang@ucas.ac.cn).}
\thanks{Sam Kwong is with the Department of Computer Science, City University of Hong Kong, Hong Kong SAR, China, and also with the City University of Hong Kong Shenzhen Research Institute, Shenzhen 51800, China (e-mail: cssamk@cityu.edu.hk).}

}

\markboth{IEEE TRANSACTIONS ON CYBERNETICS}
{Shell \MakeLowercase{\textit{et al.}}: Bare Demo of IEEEtran.cls for IEEE Journals}
\maketitle

\begin{abstract}
The goal of co-salient object detection (CoSOD) is to discover salient objects that commonly appear in a query group containing two or more relevant images. Therefore, how to effectively extract inter-image correspondence is crucial for the CoSOD task. In this paper, we propose a global-and-local collaborative learning architecture, which includes a global correspondence modeling (GCM) and a local correspondence modeling (LCM) to capture comprehensive inter-image corresponding relationship among different images from the global and local perspectives. Firstly, we treat different images as different time slices and use 3D convolution to integrate all intra features intuitively, which can more fully extract the global group semantics. Secondly, we design a pairwise correlation transformation (PCT) to explore similarity correspondence between pairwise images and combine the multiple local pairwise correspondences to generate the local inter-image relationship. Thirdly, the inter-image relationships of the GCM and LCM are integrated through a global-and-local correspondence aggregation (GLA) module to explore more comprehensive inter-image collaboration cues. Finally, the intra- and inter-features are adaptively integrated by an intra-and-inter weighting fusion (AEWF) module to learn co-saliency features and predict the co-saliency map. The proposed GLNet is evaluated on three prevailing CoSOD benchmark datasets, demonstrating that our model trained on a small dataset (about 3k images) still outperforms eleven state-of-the-art competitors trained on some large datasets (about 8k\textasciitilde200k~images).
\end{abstract}


\IEEEpeerreviewmaketitle

\section{Introduction} \label{sec1}

\IEEEPARstart{S}{alient} object detection (SOD) simulates the human visual attention mechanism to locate the visually attractive or most prominent objects/regions from a scene \cite{crm2019tcsvt}, which has been applied to a large number of vision tasks, such as image classification \cite{01guo2019visual}, image compression \cite{02han2006image}, image retargeting \cite{032014Physical}, \etc. With different data inputs, the SOD task can be further divided into some sub-tasks, such as RGB-D SOD \cite{crm-acmmm,crm-tip,crm2016spl,crm2020tcgoing,crmDPANet,crm21acmmmRGBD,crm20eccv,crmASIF,crm-tmm}, video SOD \cite{crm2019tipVideo}, remote sensing SOD \cite{crm2019tgrs,crmDAFNet,crm2021tgrs,9756846,crm-nc}, \etc.
In addition, with the explosive growth of data volume in recent years, people sometimes need to collaboratively perceive multiple relevant images, such as identifying teammates wearing the same uniform in different images. Therefore, simulating the human co-processing capabilities, co-salient object detection (CoSOD) aims to localize the common and salient objects in an image group containing multiple relevant images. As can be seen from the definition of CoSOD, it contains two key matters: the saliency attribute and the repetitiveness attribute. In other words, in addition to capturing the saliency attribute of each image, the interactive relationships among different images play an important role in determining whether the objects are shared in the entire image group. Moreover, the co-salient objects only belong to the same semantic category, but their appearance characteristics vary differently, which undoubtedly increases the difficulty of the task.

The existing CoSOD methods can be roughly divided into traditional methods \cite{CCS,cong2017,icme18,crm2019tmm,crm2018tip} and deep-learning-based methods \cite{GCS,RCGS,CoADNet,GCoNet,GICD,GCAGC,MGL,MCFIP,SAFS,GoNet,RCNet,ICNet,Ren0LZBS20,TPAMI21,CoSformer,GAN,iccv21}. Traditional methods usually use hand-crafted features to model the inter-image relationships into feature clustering, similarity matching, rank constraint, \etc.
However, their performance is limited due to the lack of high-level semantic representation, especially for the complex scenes. Since entering the deep learning era, the CoSOD task has achieved rapid development, and its performance has continued to set new records. For the process of generating group semantics, the modeling of the relationship among images can be classified into three categories: One is the global model, which combines all the intra features into the designed model to learn the global semantics \cite{GCS,RCGS,CoADNet,GCoNet,GICD,GCAGC}. Another is the local model, which decouples the inter-image relationship into the multiple pairwise correspondences \cite{ICNet}. The last one is the recurrent model, which utilizes recursive structure (such as RNN or GRU) to capture the group semantics \cite{RCNet,Ren0LZBS20}. Among them, the recurrent model is sensitive to the input order of the images, the global model is closer to a blind processing method requiring a delicately designed network to capture better group semantics, and the local model has a clearer physical meaning, but are more computationally expensive. On the whole, the global model and local model are naturally complementary to a certain degree, but there is no work to integrate them organically under a unified framework.

\begin{figure}[!t]
	\centering
	\centerline{\includegraphics[width=0.5\textwidth]{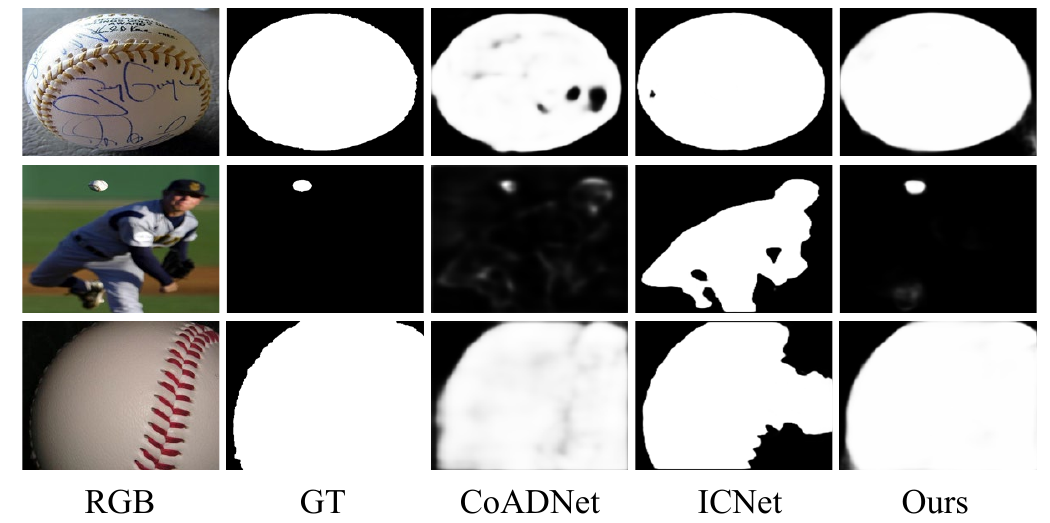}}
	\caption{Visual examples of different CoSOD models. CoADNet \cite{CoADNet} and ICNet \cite{ICNet} are the global modeling  and local modeling methods of extracting inter-image correspondence, respectively. Our method shown in the last column utilizes a global and local collaborative extraction strategy.}
	\label{fig:intro}
\end{figure}

To this end, we propose a novel network based on global-and-local collaborative learning (GLNet) to fully explore the semantic association among the group images from the global and local perspectives. The inter-image relationship extraction includes a global correspondence modeling that directly integrates all intra features to learn the global semantics, and a local correspondence modeling that combines corresponding relationships between pairwise images in the group. Considering that the global modeling method directly extracts relationships of multiple images at a time, it includes the dimensions of different images other than the traditional spatial and channel dimensions. Simple 2D convolution may be difficult to extract the corresponding relationships in an image group. Thus, we introduce 3D convolution into the CoSOD task for the first time to capture a more accurate and comprehensive global inter-image relationship. For the local correspondence modeling, we design a pairwise correlation transformation (PCT) to explore similarity correspondence between pairwise images. The differences with the existing local model \cite{ICNet} are reflected in two aspects: (1) When measuring the pairwise correlation, the PCT is executed at the pixel level. Each spatial location is associated with all other locations through a max-pooling layer, which has the global description capability. (2) When fusing the multiple local correspondences, we use progressive 3D convolution layers, which assist in reducing the redundancy and suppressing the interference. In addition, the global and local inter information are integrated through a global-and-local correspondence aggregation module, and an intra-and-inter weighting fusion strategy is designed to adaptively integrate the intra and inter features into co-saliency features to predict the corresponding co-saliency map. As can be seen in Fig. \ref{fig:intro}, our method can accurately and completely detect the co-salient object, even if the size of the object varies dramatically.

The major contributions of the proposed method are summarized as follows:
\begin{itemize}
	\item[1)]
	We propose an end-to-end network for co-salient object detection, the core of which is to capture a more comprehensive inter-image relationship through the global-and-local collaborative learning. The global correspondence modeling module directly extracts the interactive information of multiple images intuitively, and the local correspondence modeling module defines the inter-image relationship through the form of multiple pairwise images.
\end{itemize}
\begin{itemize}
	\item[2)]
	For the global correspondence modeling, different images are regarded as different time dimensions, and thus 3D convolution is used instead of the usual 2D convolution to capture global group semantics.
\end{itemize}
\begin{itemize}
	\item[3)]
	For the local correspondence modeling, we design a pairwise correlation transformation to explore similarity correspondence between pairwise images. The two modules work together to learn more in-depth and comprehensive inter-image relationships.
\end{itemize}

The rest of this paper is organized as follows. Section \ref{sec2} reviews the related works. Section \ref{sec3} introduces the proposed GLNet method in detail. The experimental results with quantitative evaluation are presented in Section \ref{sec4}. Finally, the conclusion is drawn in Section \ref{sec5}.

\section{Related Work} \label{sec2}
In this section, we discuss and summarize related work on salient object detection and co-salient object detection.
\begin{figure*}[!ht]
	\includegraphics[width=1.01\textwidth]{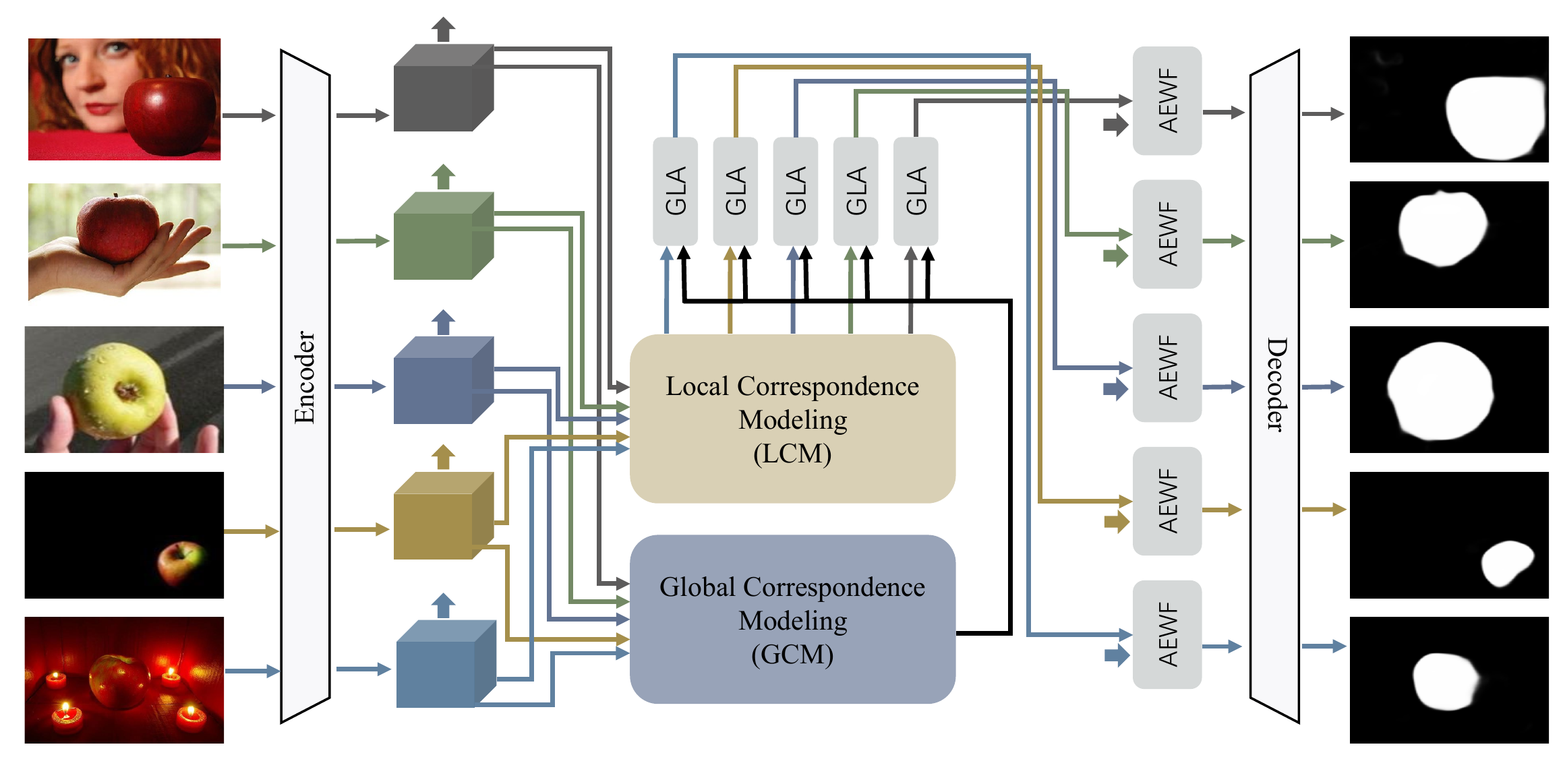}
	\caption{The flowchart of the proposed GLNet. Given a query image group, we first obtain the deep intra features with a shared backbone feature extractor. A global correspondence modeling (GCM) module is applied to capture the global interactive information among multiple images, while a local correspondence modeling (LCM) module further collects the local interactive information between pairwise images. The global and local features are generated into inter features through a global-and-local correspondence aggregation (GLA) module. Then, the intra and inter features are further adaptively aggregated through an intra-and-inter weighting fusion (AEWF) module to learn co-saliency features. In the end, a multi-layer deconvolution is used to predict the full-resolution co-saliency maps $M$.}
	\label{framwork}
\end{figure*}

Salient object detection has attracted the attention of many researchers, and a large number of methods have been proposed \cite{RC,DSR,SMD,RRWR,DSG,R6}. Li \etal \cite{DSR} used reconstruction error to measure the significance of a region, and the reconstruction error corresponding to the significant region was larger. Li \etal \cite{RRWR} proposed a novel graph-based saliency detection algorithm, which formulated the pixel-wised saliency maps using the regularized random walks ranking. Wang \etal \cite{R6} presented a supervised framework for saliency detection, which combines a set of low-, mid-, and high-level features to predict the possibilities of each region being salient. Recently, deep learning has proven its power in salient object detection, and its performance is constantly improving \cite{NCCB,DCL,zhang2020few,zhang2019synthesizing,DSS,F3Net,RADF,UCF,RFCN,EGNet,DNA2021,DNMF}. Wei \etal \cite{F3Net} selectively integrated features at different levels and refined multi-level features iteratively with feedback mechanisms to predict salient object map. Zhao \etal \cite{EGNet} used the contour as another supervision information to guide the learning process of bottom features in salient object detection.

Due to the lack of constraints on the relationship between images, directly transplanting the SOD model to CoSOD task often fails to achieve the desired effect. Therefore, establishing the inter-image constraint relationships are the key to solving this task. Originally, hand-crafted features (such as color, texture, and contrast) are used to model the inter-image constraint relationships in the CoSOD task \cite{CCS,cong2017,icme18,crm2019tmm,crm2018tip}. However, the expression ability of hand-crafted features is very limited, especially for the objects with large appearance variations and complex background textures, resulting in unsatisfactory prediction result.
Recently, deep learning-based methods have helped the CoSOD task achieve considerable performance gains \cite{MGL,MCFIP,SAFS,GoNet,RCNet,ICNet,GCS,RCGS,CoADNet,GCoNet,GICD,GCAGC,Ren0LZBS20,TPAMI21,CoSformer,GAN,iccv21}. Wei \etal \cite{GCS} proposed the first CNNs based co-salient object detection model, where the individual image saliency features were simply concatenated to learn the group-wise feature representation. Wang \etal \cite{RCGS} learned the group-wise semantic vector to represent the inter-image correspondence. Zhang \etal \cite{CoADNet} designed an effective aggregation-and-distribution strategy for inter-image modeling and achieved highly competitive performance.
Zhang \etal \cite{GICD} proposed a gradient induction model that used image gradient information to draw more attention to the discriminative common saliency features. Zhang \etal \cite{GCAGC} designed an adaptive graph convolutional network to capture the intra- and inter- image correspondence of an image group. Jin \etal \cite{ICNet} proposed that the weighted average each pairwise image was masked by the predicted saliency map as group semantics, and then used to compare the cosine similarity of each position of each image. Fan \etal \cite{TPAMI21} learned common information using co-attentional projection strategy, and at the same time established a CoSOD3k dataset for the CoSOD task. Tang \etal \cite{CoSformer} regarded the CoSOD task as an end-to-end sequence prediction problem and proposed the co-salient object detection transformer (CoSformer) network. Qian \etal \cite{GAN} used a two-stream encoder generative adversarial network (TSE-GAN) with progressive training. Zhang \etal \cite{iccv21} proposed a consensus-aware dynamic convolution model to perform the summarize and search process.

For the process of generating group semantics, these methods perceive the inter-image relationships from a global perspective or a local perspective. The global perspective uses a direct way to extract the group semantic features of multiple images at once, while the local perspective allows the network to learn pairwise relationships in a step-by-step manner and finally fuse them into group semantics. In fact, we can capture group semantics in an all-round way from two perspectives. Therefore, we propose a novel global-and-local collaborative learning network (GLNet), where the GCM and LCM modules are designed to fully explore the semantic association among the group image from different perspectives.

\begin{figure*}[!t]
	\centering
	\centerline{\includegraphics[width=1.02\textwidth]{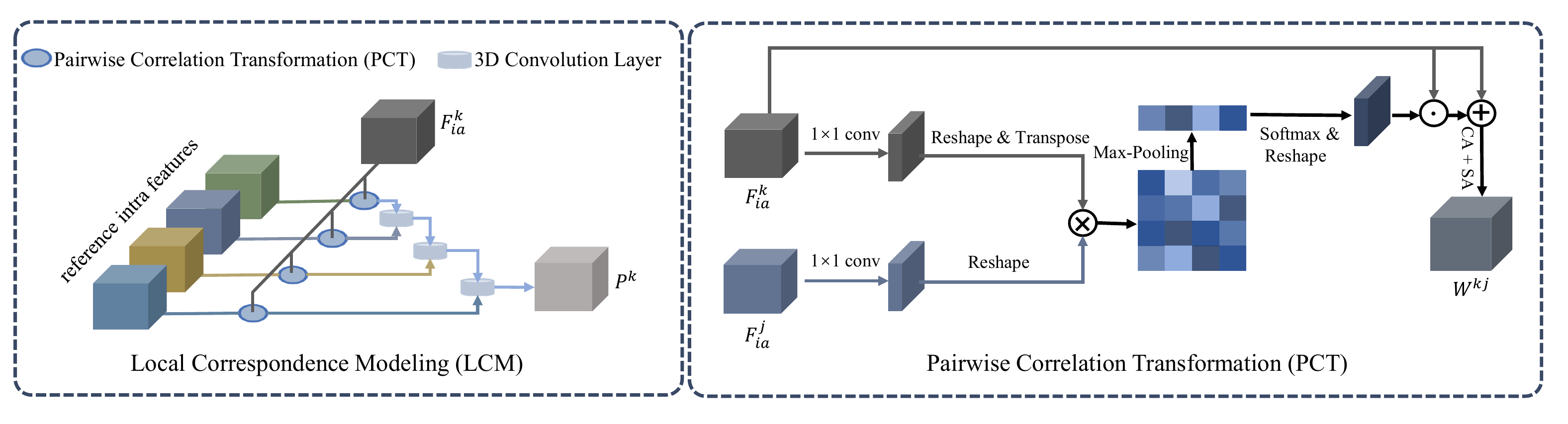}}
	\caption{Left: Local correspondence modeling (LCM). Just for intuition, here we show the calculate of the local inter features $P^k$ for one image $I^k$ from a group of 5 images. Right: Pairwise correlation transformation (PCT). $\otimes$ and $\oplus$ denote the matrix multiplication and element-wise addition, respectively. $\odot$ means element-wise multiplication broadcasted along feature planes. In fact, we respectively perform pairwise correlation transformation on the current image $I^k$ and the other reference images $I^j$ in the group to generate multiple local inter features $W^{kj}$.}
	\label{fig:LIFE}
\end{figure*}
\section{Proposed Method} \label{sec3}

\subsection{Overview}
The framework of the proposed method is shown in Fig. \ref{framwork}. Given a query group including $N$ relevant images $I={\{I^n\}}_{n=1}^N$, the goal of the CoSOD task is to detect commonly salient objects and predict the corresponding co-saliency maps $M={\{M^n\}}_{n=1}^N$. First, we use the backbone network (\eg, VGG \cite{vgg}, ResNet\cite{ResNet}, Dilated ResNet \cite{DR}) to extract the multi-level features of each image in the group. Considering that the top-level convolutional features contain more semantic information and are more suitable for exploring the corresponding information among the group images, we use them as the final intra features $F^n_{ia}\in\mathbb{R}^{C\times H\times W}$, where $C$, $H$, $W$ denote the channel, height, and width of the feature map, respectively.
Then, in order to fully exploit the inter-image cues among different images, a global-and-local collaborative learning architecture is designed.
In addition to the designed global correspondence modeling (GCM) that directly fuses all the intra features through the 3D convolution, we also design a local correspondence modeling (LCM) method from a local perspective by decoupling the multi-image relationship into multiple pairwise-image correspondences. Furthermore, the interactive information of these two complementary aspects is integrated through a global-local aggregation (GLA) module to learn the collaborative cues among the group images. Finally, the intra and inter features are adaptively integrated through an intra-and-inter weighting fusion (AEWF) module and are fed into a decoder network to predict the corresponding co-saliency maps $M$.

\subsection{Global Correspondence Modeling}
Intuitively, the co-salient objects in an image group have the same semantic attributes in features. The extraction of the inter-image correspondence among images is essential to determine which features are shared by multiple images. Therefore, the most intuitive way to achieve this is to directly integrate the different intra features in the group, and then learn the common relationship attributes through some 2D convolution layers \cite{GCS,RCGS}. This is a global modeling method that can directly extract multiple-image relationships at once. However, it is very difficult to adequately extract the relationship between different images using traditional 2D convolutional layers because a new dimension of different images is added compared to a single image (we will verify this in the ablation study of Section \ref{sec4}). The superiority of 3D convolution has been previously demonstrated in other vision tasks \cite{R5,huang2021scribble}. As stated in \cite{R5}, replacing 2D convolutions with 3D convolutions not only reduces the size of the model, but also improves the performance of the model.
Inspired by this, we treat the different images as different time slices and use 3D convolution to replace 2D convolution for global correspondence modeling.

Specifically, the group features $\overline F\in\mathbb{R}^{C\times N \times H\times W}$ are firstly obtained by concatenating all intra features $F^n_{ia}$ of different images along channel dimension, which are the inputs of the 3D convolutional layer. Then, three 3D convolutional layers are used to capture the global correspondence:
\begin{equation}\label{equ_2}
g=\delta \circ f_{2\times 3\times 3}\circ \delta \circ f_{3\times 3\times 3}\circ \delta \circ f_{2\times 3\times 3}\circ (\overline F),
\end{equation}
where $f_{2\times 3\times 3}$ and $f_{3\times 3\times 3}$ denote 3D convolutional layers with the filter sizes of $2\times 3\times 3$ and $3\times 3\times 3$, $\delta$ is the ReLU activation function, and $\circ$ denotes function composition. In order to generate more compact global inter-image presentation, we introduce the channel attention (CA) \cite{212017Squeeze} and spatial attention (SA) \cite{202018CBAM} to highlight the important channels and spatial locations, thereby producing the final global correspondence $G\in\mathbb{R}^{C\times H\times W}$:
\begin{equation}\label{equ_1}
G=SA(CA(g)),
\end{equation}
where $CA(\cdot)$ and $SA(\cdot)$ are the channel attention operation and spatial attention operation, respectively.

\subsection{Local Correspondence Modeling}
The global correspondence modeling captures the inter-image relationship straightforwardly and globally from the image group at one time, which means that it needs to analyze the relationship between all images in the group at the same time. In fact, we can further decompose the relationship modeling among multiple images into more basic and smaller units, that is, the combination of multiple pairwise correspondences. Specifically, in order to capture the corresponding relationship between a certain image and all other $N-1$ reference images in the group, we can model the pairwise relationship between it and others separately, and then integrate these relationships together to obtain its final inter-image correspondence. This is the local correspondence modeling (LCM) we proposed, which is conducive to capturing more in-depth local interaction information among images.
In order to illustrate simply and clearly how our LCM works in the GLNet, we take the $k$-th image $I^k$ in a group including $N$ images as an example, where $k\in\{1,2,...,N\}$. The left side of Fig. \ref{fig:LIFE} shows the pipeline of LCM for calculating the local inter-image correspondence $P^k$ of image $I^k$.

Because the local correspondence modeling only needs to process two images in the group, it is different from the global correspondence model which is similar to a black-box calculation. Therefore, we design a more physically meaningful pairwise correlation method to measure the semantic correlation. We respectively calculate the correlation between the current intra features $F^k_{ia}$ and other reference intra features $F^j_{ia}$ ($j\neq k$), thereby generating the multiple pairwise local correlation features $W^{kj}\in\mathbb{R}^{C\times H\times W}$. Then, these features $W^{kj}$ regarding image $I^k$ are integrated by progressive 3D convolution layers to obtain its local inter-image correspondence $P^k\in\mathbb{R}^{C\times H\times W}$.

The CoSOD task needs to determine the common attributes of objects by modeling the correspondence among images, and further select the common and salient objects. For a pair of images, the extraction of the inter-image relationship becomes simpler and clearer, which can be defined as the correlation between two image features. If the correlation of a location is high, it means that the location is more likely to have a common object.
Specifically, we design a pairwise correlation transformation (PCT) to explore similarity correspondence between pairwise images, as shown in the right side of Fig. \ref{fig:LIFE}. Taking the current intra features $F^k_{ia}\in\mathbb{R}^{C\times H\times W}$ and reference intra features $F^j_{ia}\in\mathbb{R}^{C\times H\times W}$ as the inputs of the PCT, we first transform each of them into a new feature subspace through a $1\times 1$ convolution layer, respectively. Next, a matrix multiplication operation is performed for computing the affinity matrix $A^{kj}\in\mathbb{R}^{HW\times HW}$:
\begin{equation}\label{equ_3}
\begin{aligned}
A^{kj}=\Re_{1}(f_{1\times 1}(F^k_{ia}))^T\otimes \Re_{1}(f_{1\times 1}(F^j_{ia})),
\end{aligned}
\end{equation}
where $f_{1\times 1}$ is the 2D convolutional layer with the filter size of $1\times 1$, $\otimes$ represents the matrix multiplication, superscript $T$ denotes the transposition, and $\Re_{1}(\cdot)$ reshapes the features into the dimension of $\mathbb{R}^{C\times HW}$.
The element $a^{kj}_{pq}$ in affinity matrix $A^{kj}$ represents the correlation probability between the location $p$ in features $F_{ia}^k$ and location $q$ in features $F_{ia}^j$, and a large response corresponds to a strong correlation. Further, we need to measure the global correlation of each spatial position in image $I^k$ relative to all positions in image $I^j$. To this end, we apply the max-pooling on the affinity matrix $A^{kj}$ by row to obtain an affinity vector $\overline {A}^{kj}\in\mathbb{R}^{HW\times 1}$, then normalize and reshape it into a location-correlation weight map $\widetilde {A}^{kj}\in\mathbb{R}^{1\times H\times W}$:
\begin{equation}\label{equ_4}
\begin{aligned}
\widetilde {A}^{kj}=\Re_{2}(softmax(maxpool({A}^{kj}))),
\end{aligned}
\end{equation}
where $maxpool(\cdot)$ is the max-pooling along the row, $softmax(\cdot)$ denotes the softmax layer for feature normalization, and $\Re_{2}(\cdot)$ reshapes the features into the dimension of $\mathbb{R}^{1\times H\times W}$.

Finally, we combine the location-correlation weight map $\widetilde {A}^{kj}$ with the original features $F_{ia}^k$ in a residual connection manner, and use channel attention (CA) \cite{212017Squeeze} and spatial attention (SA) \cite{202018CBAM} for feature enhancement, thereby obtaining the pairwise correlation features $W^{kj}$ of the features $F_{ia}^k$ relative to the features $F_{ia}^j$:
\begin{equation}\label{equ_5}
\begin{aligned}
W^{kj}=SA(CA(\widetilde {A}^{kj} \odot F^k_{ia}\oplus F^k_{ia})),
\end{aligned}
\end{equation}
where $\odot$ means element-wise multiplication broadcasted along feature planes, $CA(\cdot)$ and $SA(\cdot)$ are the channel attention and spatial attention, respectively.

Repeating the above procedures, we can obtain $N-1$ pairwise correlation features $W^{kj}$ ($j\in\{1,2,...,N\}$ and $j\neq k$) of image $I^k$. Then, they are fully integrated by progressive 3D convolution layers with the filter size of $2\times 3\times 3$ to obtain the local inter features $P^k$ of image $I^k$.

\subsection{Global-and-Local Correspondence Aggregation}
\vspace{0.3cm}
Both the GCM module and the LCM module can extract the inter-image relationship among images, but their difference is that the global inter features $G$ describe the global correspondence among multiple images in the group, while the local inter features $P=\{P^k\}_{k=1}^N$ obtain the local relationship from multiple pairwise images. These two modules define the inter-image relationship from different angles, and there is a certain degree of complementarity. Thus, we combine them to learn more comprehensive inter features $F_{ie}^k\in\mathbb{R}^{C\times H\times W}$ of image $I^k$ through a global-and-local correspondence aggregation (GLA) module. Concretely, taking into account the advantages of 3D convolution, we still use it for fusion learning. The GLA consists of a 3D convolutional layer with the filter size of $2\times 3\times 3$, a channel attention scheme, and a spatial attention scheme. The global inter features $G\in\mathbb{R}^{C\times H\times W}$ and local inter features $P^k\in\mathbb{R}^{C\times H\times W}$ are first concatenated along the channel dimension as the inputs of GLA module. Then, we can obtain the final inter features $F_{ie}^k$ via a 3D convolutional layer with ReLU activation as follows:
\begin{equation}\label{equ_6}
\begin{aligned}
F_{ie}^k=SA(CA(\delta(f_{2\times 3\times 3}([G,P^k])))),
\end{aligned}
\end{equation}
where $[\cdot,\cdot]$ denotes the concatenation along the channel dimension.

\subsection{Intra-and-Inter Weighting Fusion}
\vspace{0.3cm}
As described previously, we construct the inter features $F_{ie}^k$ from global and local perspectives, aiming to capture the corresponding relationships among group images. Returning to the definition of co-salient object detection, we need to consider both the saliency attributes within an image and the correspondence among different images. Therefore, we should combine intra features and inter features to obtain the co-saliency features. But for different scenes, intra features and inter features play different roles, so we design a dynamic weighting strategy to adaptively fuse them instead of the clumsy addition or concatenation fusion strategies. Concretely, we firstly concatenate the features $F_{ie}^k$ and $F_{ia}^k$, and apply a $1\times 1$ convolutional layer for channel reduction to generate $F^k_{cat}\in\mathbb{R}^{C\times H\times W}$. Then, we learn a weighting map $\alpha^k \in\mathbb{R}^{C\times H\times W}$ through a bottleneck convolutional layer, which determines the importance between intra and inter features. Therefore, the final co-saliency features $F^k_{co}$ of image $I^k$ can be obtained by:
\begin{equation}\label{equ_7}
\begin{aligned}
F^k_{co}=\alpha^k \odot F_{ie}^k+(1-\alpha^k)\odot F_{ia}^k,
\end{aligned}
\end{equation}
\begin{equation}\label{equ_8}
\begin{aligned}
\alpha^k=\sigma(f_p(CA(F^k_{cat}))),
\end{aligned}
\end{equation}
where $f_p$ is a bottleneck convolutional layer, and $\sigma$ represents the sigmoid activation. For the final co-saliency map $M^k$ prediction, we need to up-sample the co-saliency features $F^k_{co}$ to the same resolution with the input image, for which multi-layer deconvolution operations are adopted.

\subsection{Loss function}
\vspace{0.3cm}
During training, the proposed network is optimized by using the binary cross-entropy loss between each co-saliency map $M^n$ and the ground truth $T^n$ in an end-to-end manner. Specifically, given $N$ co-saliency maps and ground-truth masks (\ie, $\{M^n\}_{n=1}^N$ and $\{T^n\}_{n=1}^N$) in an image group, we define the loss function $\ell$ by calculating the binary cross-entropy loss over all the samples:

\begin{equation}
\label{equ_9}
\begin{aligned}
\ell=\frac{1}{N}\sum_{n=1}^{N}\ell_{bce}^n,
\end{aligned}
\end{equation}
where $\ell_{bce}^n=-[T^nlog(M^n)+(1-T^n)log(1-M^n)]$ is the binary cross-entropy loss of the sample $n$, and $N$ denotes the image number in an image group.
\begin{figure*}[!ht]
	\includegraphics[width=1.01\textwidth]{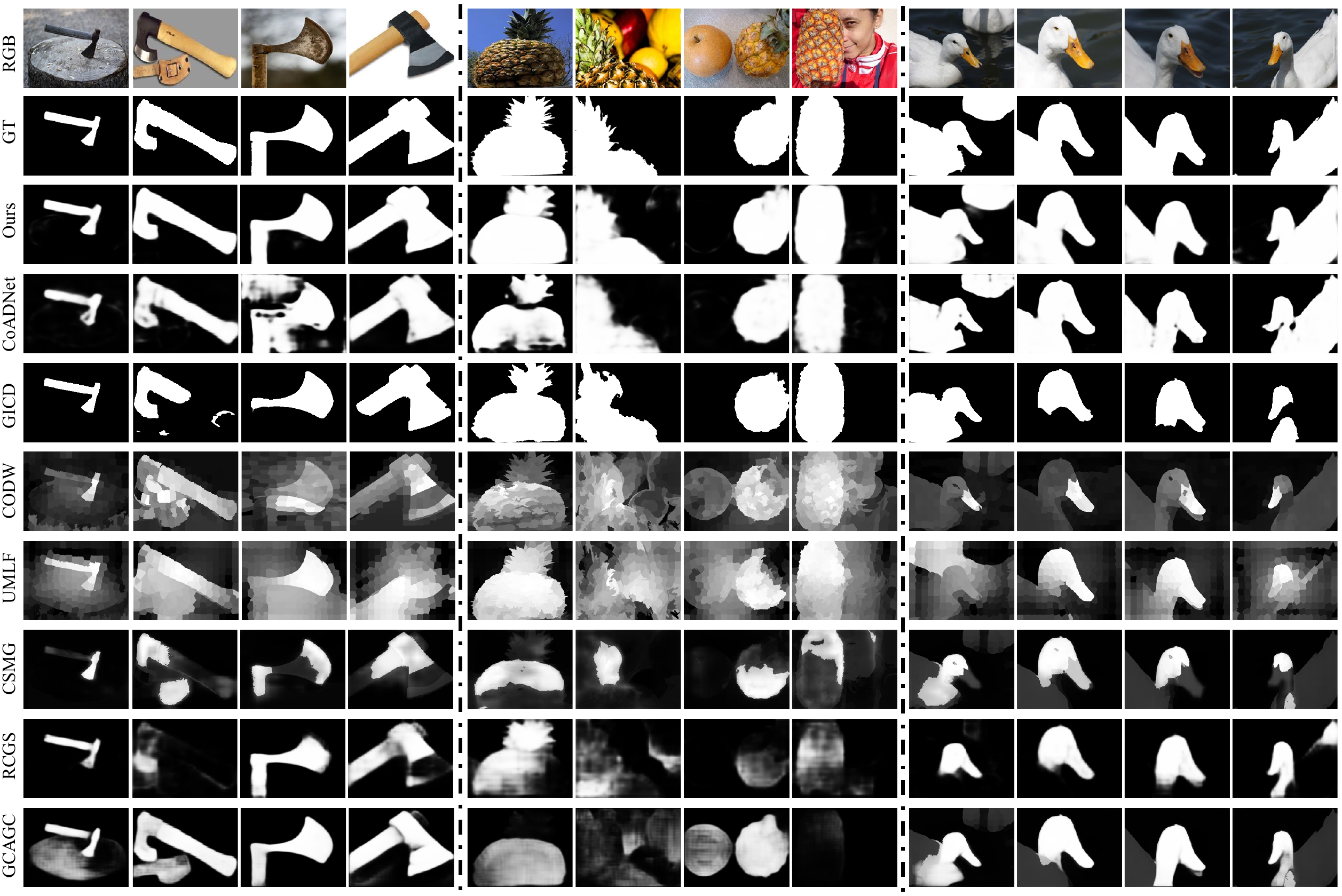}
	\caption{Visual comparisons of different methods. The first image group is axe, the second one is the pineapple, and the last one is white goose.}
	\label{keshihua}
\end{figure*}

\begin{figure*}[!ht]
	\includegraphics[width=1\textwidth]{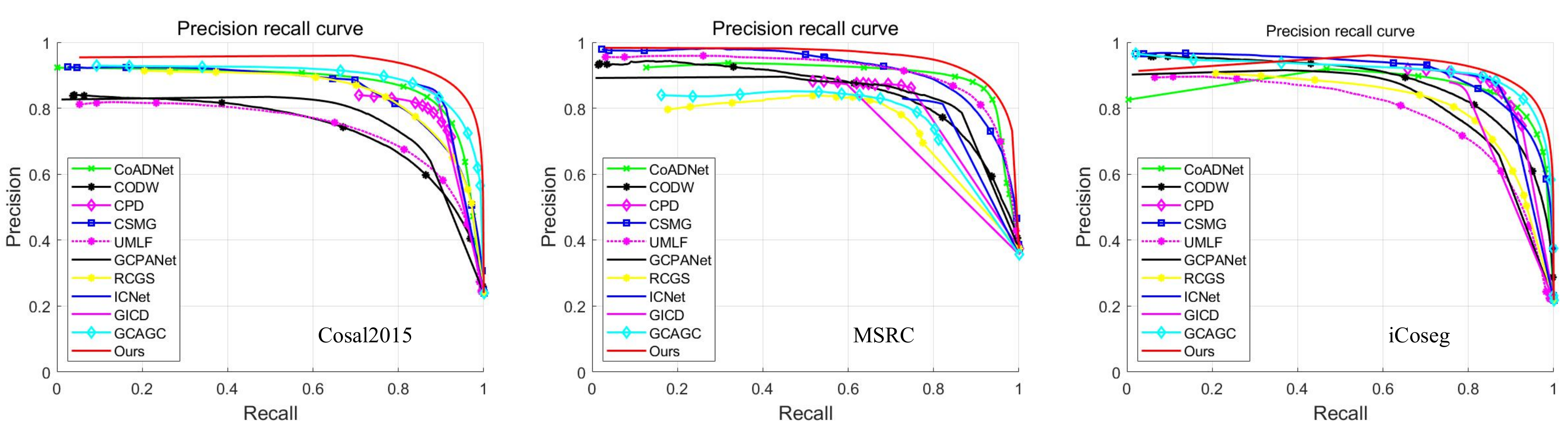}
	\caption{P-R curves of different methods on three benchmark datasets.}
	\label{PR}
\end{figure*}

\begin{table*}
	\centering
	\caption{Quantitative comparisons of our GLNet with other state-of-the-arts in terms of max F-measure ($F_\beta$), S-measure ($S_\alpha$), max E-measure ($E_\varphi$) and MAE score on three benchmark datasets. ``Co'' denotes the CoSOD model, and ``Sin'' represents the single-image SOD model. ``$\uparrow$'' (``$\downarrow$'') means that higher (lower) is better. ``-S'' means the subset of the correspond dataset. ``Num.'' represents the number of training images. Black bold fonts indicates the best performance.}
	\begin{center}
		\renewcommand{\arraystretch}{1.65}
		\setlength{\tabcolsep}{0.4mm}{
			\begin{tabular}{c|c|c|c|c|cccc||cccc||cccc}
				\hline\hline
				\multirow{2}{*}{Method} & \multirow{2}{*}{Pub. \& Year}   & \multirow{2}{*}{Type} & \multirow{2}{*}{Training Set} & \multirow{2}{*}{Num.} & \multicolumn{4}{c||}{Cosal2015}                      & \multicolumn{4}{c||}{MSRC}                           & \multicolumn{4}{c}{iCoseg}                          \\ \cline{6-17}
				&                                 &                       &                           &                               & $S_\alpha\uparrow$          & $F_\beta\uparrow$     & $E_\varphi\uparrow$    & MAE $\downarrow$        & $S_\alpha\uparrow$         & $F_\beta\uparrow$      & $E_\varphi\uparrow$    & MAE $\downarrow$      & $S_\alpha\uparrow$       & $F_\beta\uparrow$    & $E_\varphi\uparrow$     & MAE $\downarrow$                   \\ \hline
				CPD \cite{CPD}                     & \multicolumn{1}{c|}{CVPR'19}  & Sin                   & ---                       & ---                           & .8168          & .7868         & .8438 & .0976          & .7039          & .7731       & .7942   & .1818          & .8565          & .8451   & .8933       & .0579          \\
				GCPANet \cite{GCPANet}                 & \multicolumn{1}{c|}{AAAI'20}  & Sin                   & ---                       & ---                           & .7668          & .7077      & .8000    & .1085          & .7312          & .7973      &.8372    & .1630          & .7718          & .7344      & .8260    & .0983          \\ \hline
				CODW \cite{CODW}                    & \multicolumn{1}{c|}{CVPR'15}  & Co                 & ImageNet \cite{imagenet} pre-train     & ---                                 & .6501          & .6715      & .7524    & .2741          & .7111          & .7971      &.8204    & .2683          & .7510          & .7857     & .8355     & .1782          \\
				UMLF \cite{UMLF}                    & \multicolumn{1}{c|}{TCSVT'18} & Co                & Cosal2015 \cite{CODW}+MSRC-V1 \cite{MSRC}     & 2,135                      & .6649          & .6956      & .7723    & .2691          & .8057          & .8627     & .8828     & .1795          & .6828          & .7262      & .7997    & .2389          \\
				CSMG \cite{CSMG}                    & \multicolumn{1}{c|}{CVPR'19}  & Co              & MSRA-B \cite{MSRA-B}      & 2,500                                             & .7757          & .7869      & .8435   & .1309          & .7091          & .8565      & .8603    & .2008          & .8122          & .8371      & .8851    & .1050          \\
				RCGS \cite{RCGS}                    & \multicolumn{1}{c|}{AAAI'19}  & Co                   & COCO-SEG \cite{RCGS}   & 200,000                                      & .7958          & .7790      & .8529    & .1005          & .6631          & .6921     & .7191     & .2193          & .7860          & .7273      & .8177    & .0976          \\
				GICD \cite{GICD}                    & \multicolumn{1}{c|}{ECCV'20}  & Co                 & DUTS-S  \cite{DUTS}   & 8,250                                                  & .8389          & .8354     & .8822     & .0730          & .6728          & .6931     & .7309     & .1927          & .8198          & .8244     & .8841     & .0695          \\
				GCAGC \cite{GCAGC}                   & \multicolumn{1}{c|}{CVPR'20}  & Co                & COCO-SEG \cite{RCGS}+MSRA-B \cite{MSRA-B}      & 202,500                                       & .8433          & .8482          & .9003   & .0792          & .6768          & .7086     & .7424     & .2073          & .8606          & .8622     & .9098     & .0773          \\
				ICNet \cite{ICNet}                   & \multicolumn{1}{c|}{NeurIPS'20}  & Co                  & COCO-S  \cite{coco}    & 9,213                                                       & \textbf{.8487} & .8456          & .8917   & .0667          & .7359          & .8048     & .8192     & .1587          & .8563          & .8625    & .9150      & .0509          \\
				CoADNet \cite{CoADNet}                 & \multicolumn{1}{c|}{NeurIPS'20}  & Co                  & CoSOD3k \cite{CoSOD3k}+DUTS \cite{DUTS}   & 18,888                                          & .8454          & .8592      & .9066    & .0818          & .7670          & .8347     & .8941     & .1558          & .8569          & .8784     & .9272     & .0725          \\
				GCoNet \cite{GCoNet}                   & \multicolumn{1}{c|}{CVPR'21}  & Co                     & DUTS-S \cite{DUTS}   & 8,250                                           & .8454          & .8474     & .8881     & .0680          & .6681              & .7174        & .7402      & .1867             & .8333              & .8340        & .8842      & .0677             \\
				GLNet                   & \multicolumn{1}{c|}{---}         & Co                & CoSOD3k \cite{CoSOD3k}      & 3,316                                           & .8460          & \textbf{.8936}  & \textbf{.9377} & \textbf{.0648}   & \textbf{.8620}  & \textbf{.8945} & \textbf{.9015} & \textbf{.0956} & \textbf{.8629}  & \textbf{.8880} & \textbf{.9483} & \textbf{.0500} \\\hline\hline
		\end{tabular}}
	\end{center}
	\label{table:1}
\end{table*}

\section{Experiments} \label{sec4}
In this section, we first introduce the datasets, the evaluation, and the implementation details. Then, we make some comparisons between the proposed GLNet and the state-of-the-art (SOTA) SOD and CoSOD methods to demonstrate our superiority. Lastly, the ablation studies will be discussed to investigate the effectiveness of each module.

\subsection{Datasets and Evaluation Metrics}
\subsubsection{Benchmark Datasets} We evaluate different methods on three widely used CoSOD benchmark datasets, including MSRC \cite{MSRC}, iCoseg \cite{icoseg}, and Cosal2015 \cite{CODW}. Each image in these datasets has a corresponding pixel-wise CoSOD ground truth. Among them, the MSRC dataset \cite{MSRC} is the first available dataset for CoSOD and consists of 210 images in 7 groups after removing the non-saliency $grass$ group. The iCoseg dataset \cite{icoseg} contains 643 images distributed in 38 image groups, and the number of images in each group varies from 4 to 42. The Cosal2015 dataset \cite{CODW} is a large and challenging dataset containing 2015 images in 50 groups, which includes various challenging factors, such as complex background and occlusion issues.

\subsubsection{Evaluation metrics} For quantitative evaluation, we introduce five widely used evaluation indicators in the SOD scenario, including the precision-recall (P-R) curve, F-measure ($F_\beta$) \cite{Fmeasure}, MAE score \cite{TIP18}, S-measure ($S_\alpha$) \cite{S-measure}, and E-measure ($E_\varphi$) \cite{Emeasure}. The P-R curve intuitively shows the variation trend between different precision and recall scores, and the closer the curve is to (1,1), the better the method performance.
F-measure ($F_\beta$) is a comprehensive measurement of precision and recall values. The specific formula is written as follows:
\begin{equation}
\label{F}
F_{\beta}=\frac{(1+\beta^{2})precision\times recall}{\beta^{2}\times precision + recall},
\end{equation}
where $precision$ and $recall$ are the precision score and recall score, respectively, and $\beta^{2}$ is fixed to $0.3$ as suggested in \cite{Fmeasure}.

MAE score is used to calculate the pixel-wise difference between the predicted co-saliency map $M$ and ground truth $T$, which is formulized as:
\begin{equation}
\label{MAE}
MAE =\frac{1}{W\times H} \sum_{w=1}^{W} \sum_{h=1}^{H} |M(w,h)-T(w,h)|,
\end{equation}
where $W$ and $H$ are the width and height of the image, respectively.

S-measure ($S_\alpha$) describes the structural similarity between the co-saliency map and ground truth, which is defined as:
\begin{equation}
\label{Sm}
S_\alpha = \alpha \times S_o+(1-\alpha) \times S_r,
\end{equation}
where $S_o$ represents the object-aware structural similarity, summing foreground and background comparison terms with weights.
$S_r$ denotes the region-aware structural similarity, which divides the saliency map and ground-truth into multiple regions and computes the weighted summation of the corresponding structural similarity measure. $\alpha$ is the balance parameter and set to
$0.5$ as suggested in \cite{S-measure}.

E-measure ($E_\varphi$) consists of a single term to account for both pixel and image-level properties, which is computed as:
\begin{equation}
\label{E}
E_\varphi=\frac{1}{W \times H}\sum_{w=1}^{W} \sum_{h=1}^{H}\theta(\varphi),
\end{equation}
where $\theta(\varphi)$ indicates the enhanced alignment matrix.

Among these indicators, the larger the $F_\beta$, $E_\varphi$, and $S_\alpha$, the better the performance, while the MAE value is just the opposite (\ie, the smaller value corresponds to the better performance).

\subsection{Implementation Details}
\subsubsection{Training data} Most of the previous state-of-the-art CoSOD models are trained on the large-scale COCO-SEG dataset \cite{RCGS}, which contains more than 200,000 image samples.
As pointed out in \cite{CoADNet}, although there are a large amount of data in COCO-SEG, many objects in this dataset do not have the saliency attributes, so it is not completely suitable for the CoSOD task.
Following the settings in \cite{CoADNet}, we take the CoSOD3k dataset \cite{CoSOD3k} as our training dataset. The CoSOD3k dataset is specially designed for the CoSOD task, which contains 3,316 images distributed in 160 image groups.

\subsubsection{Implementation settings} We implement the proposed model via PyTorch toolbox and train it on an RTX 2080Ti GPU in an end-to-end manner. We also implement our network by using the MindSpore Lite tool \footnote{\url{https://www.mindspore.cn/}}. In order to avoid overfitting, we use random flipping and rotating to augment the training samples. Following the input settings in \cite{GCS,CoADNet}, we set the input image number of the GLNet to 5, and select mini-batch groups from all categories in the CoSOD3k dataset. Due to the limited computing resources, all input images are resized to $160 \times 160$, and rescaled to the original size for testing. We use Adam \cite{adam} to train our model with weight decay of $10^{-4}$ and the training process converges until 40,000 iterations. The learning rate decay strategy adopts cosine annealing decay, where the initial and minimum learning rate are set to $5e^{-6}$ and $5e^{-7}$, respectively.  The code and results can be found at \url{https://rmcong.github.io/proj_GLNet.html}.
\begin{figure}[!t]
	\centering
	\centerline{\includegraphics[width=0.45\textwidth]{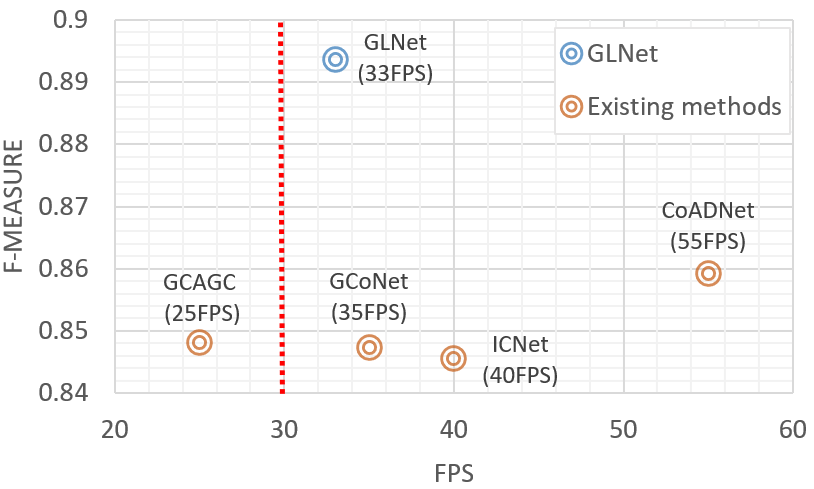}}
	\caption{Comparison of FPS and F-measure on the Cosal2015 dataset.}
	\label{Fig1}
\end{figure}
\subsubsection{Backbone setting} In our experiment, we choose the VGG16 network as our feature extractor for two reasons. Firstly, most of the previous methods used the VGG as the backbone, such as GCAGC \cite{GCAGC}, ICNet \cite{ICNet}, and so on. Therefore, for a fair comparison, we also choose the VGG network to extract the multi-level features of each image in the group, which is the dominant reason. Secondly, CoSOD task deals with image group data, using other more powerful backbone networks (\eg, Dilated ResNet \cite{DR}]) may significantly increase the model size. Taken together, after a trade-off, we chose VGG as the backbone in the experiments.

\subsection{Comparison with State-of-the-art Methods}

In order to demonstrate the effectiveness of GLNet, we compare it with eleven state-of-the-art methods on the above three datasets, including two SOD methods for single image (\eg, GCPANet \cite{GCPANet}, and CPD \cite{CPD}), and nine CoSOD methods (\eg, GICD \cite{GICD}, GCAGC \cite{GCAGC}, CODW \cite{CODW}, UMLF \cite{UMLF}, CSMG \cite{CSMG}, RCGS \cite{RCGS}, CoADNet \cite{CoADNet}, ICNet \cite{ICNet} and GCoNet \cite{GCoNet}). For fair comparison, all results are directly provided by authors or are reproduced by the public source codes under the default parameters.

\begin{figure*}[!ht]
	\includegraphics[width=1.01\textwidth]{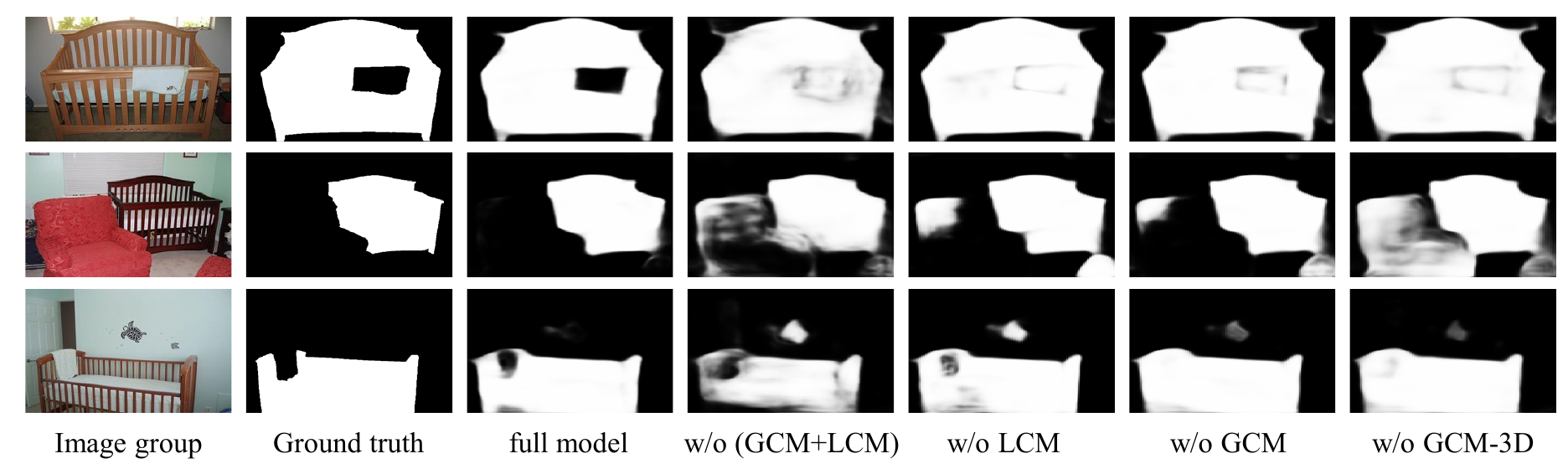}
	\caption{Visual results of ablation studies.}
	\label{duibi}
\end{figure*}

 \begin{table*}
	\centering
	\caption{Quantitative evaluation of ablation studies on the Cosal2015, MSRC and iCoseg datasets. Black bold fonts indicates the best performance.}
	\begin{center}
		\renewcommand{\arraystretch}{1.65}
		\setlength{\tabcolsep}{2.35mm}{
			\begin{tabular}{c|c|cccc||cccc||cccc}
				\hline\hline
				\multirow{2}{*}{ID}  & \multirow{2}{*}{Model} & \multicolumn{4}{c||}{Cosal2015} & \multicolumn{4}{c||}{MSRC} & \multicolumn{4}{c}{iCoseg} \\ \cline{3-14}
				&                            & $S_\alpha\uparrow$          & $F_\beta\uparrow$     & $E_\varphi\uparrow$    & MAE $\downarrow$        & $S_\alpha\uparrow$         & $F_\beta\uparrow$      & $E_\varphi\uparrow$    & MAE $\downarrow$      & $S_\alpha\uparrow$       & $F_\beta\uparrow$    & $E_\varphi\uparrow$     & MAE $\downarrow$    \\ \hline
				1                         & \emph{full model}                  & \textbf{.8460}   & \textbf{.8936}  & \textbf{.9377} & \textbf{.0648}   & \textbf{.8620}  & \textbf{.8945} & \textbf{.9015} & \textbf{.0956} & \textbf{.8629}  & \textbf{.8880} & \textbf{.9483} & \textbf{.0500} \\\hline
				2                                    & $w/o~(GCM+LCM)$                & .7937   & .8446  & .8978 & .1120   & .8104  & .8498 & .8551 & .1450 & .8326  & .8378 & .9103 & .0799 \\ \hline
				3                                        & $w/o~LCM$                    & .8283   & .8550 & .9113  & .0783   & .8500  & .8908 &  .8994 & .1031 & .8607  & .8777 & .9403 & .0552 \\ \hline
				4                        & $w/o~GCM$                    & .8295   & .8524 & .9100  & .0821   & .8536  & .8868 & .8988 & .1065 & .8618  & .8723 & .9403 & .0572 \\
				5                                             & $w/o~GCM$-$3D$                     & .8150   & .8657 &.9132  & .0977   & .8611  & .8914 & .9013 & .1079 & .8584  & .8803 & .9451 & .0628 \\
				\hline\hline
		\end{tabular}	}
	\end{center}
	\label{table:2}
\end{table*}

Fig. \ref{keshihua} shows the visual comparison results of different CoSOD methods. Intuitively, we can see that our method locates the co-salient objects more accurately and completely than other competing methods. For example, in the third group (\ie, goose), only our method can accurately locate the co-salient object in each image and show superiority in terms of the internal consistency, even if the co-salient object is partially occluded. When there are obvious appearance changes, strong semantic interference, and complex backgrounds in the scene (such as the first group of axes in Fig. \ref{keshihua}), our model can still accurately, completely, and clearly detect the co-salient object, benefiting from the comprehensive modeling of the interactive information among images.

For quantitative evaluation, Fig. \ref{PR} shows the P-R curves of all compared methods on three benchmark datasets. We can observe that our GLNet outperforms the other state-of-the-art methods on all datasets. In particular, the P-R curves are much higher than the other methods on the challenging and largest Cosal2015 dataset. Meanwhile, Table \ref{table:1} lists the quantitative results. In general, we can notice that our GLNet achieves better performance than most of comparison methods in terms of three evaluation indicators. For example, compared with the \emph{second best} method on the Cosal2015 dataset, the percentage gain of F-measure is 4.00\%, and the MAE score reaches 2.85\%. Our GLNet model achieves the best F-measure of 0.8945 and 0.8880 on the MSRC dataset and iCoseg dataset respectively, where the performance gains on the MSRC dataset and iCoseg dataset respectively reach 3.69\% and 1.09\% compared with the \emph{second best} method. For the E-measure, the minimum percentage gain against the comparison methods reaches 3.43\% on the Cosal2015 dataset, and 2.28\% on the iCoseg dataset, respectively.

We also draw the FPS-Fmeasure map in Fig. \ref{Fig1} for a better comprehensive evaluation of performance and time on the Cosal2015 dataset. Generally speaking, more than 30 FPS can be considered real-time, and our model reaches 33 FPS, which meets the real-time requirement. As shown in Fig. \ref{Fig1}, the GLNet achieves up to 5\% gain in F-measure compared with the existing models with similar speed. In summary, our method strikes a trade-off between performance and real-time capability.

\subsection{Ablation Study}

We conduct thorough ablation studies to analyze the effects of the key modules in our GLNet on three datasets, and the visualization and quantitative results are shown in Fig. \ref{duibi} and Table \ref{table:2}. The specific experimental settings are as follows:
\begin{itemize}
\item[1)]
  $full~model$ (model 1) is the full model containing all the modules proposed in this paper.
\end{itemize}
\begin{itemize}
\item[2)]
 $w/o~(GCM+LCM)$ (model 2) removes the GCM and LCM modules from the full model and degenerates the full model into a single-image SOD model.
\end{itemize}
\begin{itemize}
\item[3)]
 $w/o~LCM$ (model 3) represents that only the LCM module is removed and the GCM module is retained in the full model.
\end{itemize}
\begin{itemize}
\item[4)]
$w/o~GCM$ (model 4) means that only the GCM module is removed and the LCM module is retained in the full model.
\end{itemize}
\begin{itemize}
\item[5)]
$w/o~GCM$-$3D$ (model 5) refers to that only the 3D convolution in the GCM module of the full model is replaced by the ordinary 2D convolution, while still retaining the LCM module.
\end{itemize}

First, we remove the GCM and LCM modules, so that the full model degenerates into a single-image SOD model. As shown in the fourth column of Fig. \ref{duibi}, although it can detect the salient objects, there are a large number of backgrounds and non-common salient objects are wrongly preserved, such as the towel in the first image, the red sofa in the second image, and the sticker on the wall in the third image. Similarly, it performs poorly in various indicators. It can be seen from the Table \ref{table:2} that compared with \emph{full model} (model 1), the S-measure, F-measure, E-measure, and MAE score of $w/o~(GCM+LCM)$ (model 2) are decreased by 6.59\%, 5.80\%, 4.44\% and 42.14\% on the Cosal2015 dataset, respectively.

Then, we investigate the effects of the two ways of inter-image relationship modeling, \ie, GCM and LCM. From Fig. \ref {duibi}, we can observe that GCM and LCM have different performances in objects positioning and background suppression, such as the towel and sticker in the third image. Moreover, it can be observed from Fig. \ref{duibi} that when LCM and GCM are organically combined, the non-common salient region is effectively suppressed (\eg, towel, sofa, and sticker), thereby further improving the performance. By comparing various indicators, it can be found that the performance of using LCM or GCM alone is decreased compared to the \emph{full model} (model 1). For example, the MAE score of model $w/o~LCM$ (model 3) is decreased by 17.24\% on the Cosal2015 dataset, and the MAE score of model $w/o~GCM$ (model 4) is dropped by 10.23\% on the MSRC dataset.

Finally, let us verify the effect of 3D convolution in the GCM module. Compared with the \emph{full model} (model 1), all indicators in $w/o~GCM$-$3D$ (model 5) are obviously reduced on the three datasets, especially for the S-measure and F-measure on the Cosal2015 dataset. Specifically, the S-measure drops from 0.8460 to 0.8150, and the F-measure decreases from 0.8936 to 0.8657. It further shows that the 3D convolution used in the GCM module can more efficiently extract multiple image relations at once, with clear advantages over traditional 2D convolution.

In summary, the ablation studies have proved the effectiveness and advantages of proposed modules from both qualitative and quantitative aspects, including the LCM, GCM, and 3D convolution in GCM.

\subsection{Future Work}
In the future work, three aspects of CoSOD can be focused on. One is a large-scale and appropriate dataset might improve the accuracy. Currently, CoSOD3k is the largest one tailored for the CoSOD task, but also contains only 3316 images. Compared with other SOD tasks, there is much room for efforts in the dataset scale. This is also the focus of our future work, building a large-scale and specialized dataset to further promote the development of CoSOD research community. Another is the design of the model. Our current network uses a fixed number of image group inputs, which makes the network less flexible. Therefore, at the model design level, we will further explore the recurrent architecture to solve this problem and improve the detection performance. Meanwhile, different CNN models also have a certain impact on performance, as explored in \cite{CNNB}. This is also an important aspect of the design of the model in the future. Last but not the least, although our model reaches a real-time level of 33 FPS, it still has room for optimization. In the future, more attention can be paid to lightweight model design to better adapt to practical applications.

\section{Conclusion} \label{sec5}
In this paper, an end-to-end global-and-local collaborative learning network is proposed to locate the co-salient objects in an image group by capturing the inter-image information from both global and local perspectives. Our architecture contains two key modules for inter-image correspondence extraction, \ie, global correspondence modeling (GCM) and  local correspondence modeling (LCM). The GCM directly captures the global group semantics by using the 3D convolution, and the LCM explores the similarity correspondence between pairwise images through the designed pairwise correlation transformation. The two complement each other and jointly improve the accuracy of the modeling inter-image relationship. Experiments on three prevailing CoSOD datasets show that our GLNet outperforms eleven state-of-the-art competitors, even if our model is trained on a small dataset.

\par
\ifCLASSOPTIONcaptionsoff
  \newpage
\fi
{
\bibliographystyle{IEEEtran}
\bibliography{sample-base}
}

\end{document}